\documentclass[letter, 10 pt, conference]{ieeeconf}  

\IEEEoverridecommandlockouts                              

\usepackage{graphics} 
\usepackage{epsfig} 
\usepackage[linesnumbered,ruled]{algorithm2e}
\usepackage{subcaption}
\usepackage{float}
\usepackage{graphicx}
\usepackage{bm}
\usepackage{times}
\usepackage{balance}
\usepackage{amsmath}
\usepackage{breqn}
\usepackage{cite}
\usepackage{threeparttable}
\usepackage{multirow}
\usepackage{booktabs}
\usepackage{bigstrut}
\usepackage{makecell}
\usepackage{color,soul}
\usepackage{blindtext}
\usepackage{amssymb}
\usepackage{adjustbox}

\graphicspath{{figures/}}

\newcommand{\etal}{\textit{et al.}}

\title{\LARGE \bf
    Multi-Modal Graph Convolutional Network with Sinusoidal Encoding for Robust Human Action Segmentation
}

\author{Hao Xing$^1$,  Kai Zhe Boey$^1$, Yuankai Wu$^2$, Darius Burschka$^3$, Gordon Cheng$^1$
    \thanks{Authors are with $^1$Institute for Cognitive Systems, $^2$Chair of Media Technology, $^3$Machine Vision and Perception Group, School of Computation, Information and Technology, Technical University of Munich, Arcisstraße $21$, $80333$ Munich, Germany.   
    {\tt\small hao.xing@tum.de}, {\tt\small kaizhe.boey@tum.de}, {\tt\small yuankai.wu@tum.de}, {\tt\small burschka@cs.tum.edu}, {\tt\small gordon@tum.de}}%
}%

\begin{document}
	
	\maketitle
	\thispagestyle{empty}
	\pagestyle{empty}

    \begin{abstract}
		
    Accurate temporal segmentation of human actions is critical for intelligent robots in collaborative settings, where a precise understanding of sub-activity labels and their temporal structure is essential. However, the inherent noise in both human pose estimation and object detection often leads to over-segmentation errors, disrupting the coherence of action sequences. To address this, we propose a Multi-Modal Graph Convolutional Network (MMGCN) that integrates low-frame-rate (e.g., 1 fps) visual data with high-frame-rate (e.g., 30 fps) motion data (skeleton and object detections) to mitigate fragmentation. Our framework introduces three key contributions. First, a sinusoidal encoding strategy that maps 3D skeleton coordinates into a continuous sin-cos space to enhance spatial representation robustness. Second, a temporal graph fusion module that aligns multi-modal inputs with differing resolutions via hierarchical feature aggregation, Third, inspired by the smooth transitions inherent to human actions, we design SmoothLabelMix, a data augmentation technique that mixes input sequences and labels to generate synthetic training examples with gradual action transitions, enhancing temporal consistency in predictions and reducing over-segmentation artifacts. 
    
    Extensive experiments on the Bimanual Actions Dataset, a public benchmark for human-object interaction understanding, demonstrate that our approach outperforms state-of-the-art methods, especially in action segmentation accuracy, achieving F1@10: $94.5\%$ and F1@25: $92.8\%$.
    

    \end{abstract}

    \section{INTRODUCTION}
Human action segmentation, the task of temporally decomposing continuous activities into coherent sub-action units, is a cornerstone of intelligent robotic systems operating in collaborative environments. Beyond recognizing discrete actions, the systems must understand their temporal structure. 

Our preliminary studies~\cite{xing2024understanding} on human-robot collaboration reveal that encoder-decoder graph convolutional networks excel at parsing spatiotemporal information by representing human skeleton and object center points as graph nodes. Through attention mechanisms, these networks dynamically update the relationships between nodes, capturing the evolving interactions between human and objects. However, joint misdetections and tracking errors in skeleton data often lead to over-segmentation, fragmenting continuous actions into disjointed segments and disrupting the temporal coherence of action sequences. This limitation becomes particularly critical in human-object interaction scenarios, where both human pose estimation (skeleton) and object detections (bounding boxes) introduce noise. Object misdetections, such as missing and inaccurately localized bounding boxes, lead to inconsistencies in action segmentation.



Recent advances in multi-modal networks~\cite{duan2022revisiting, tani2024graph} have demonstrated promise in addressing noise-induced over-segmentation by integrating structured 3D motion data (e.g., skeletal trajectories and object centroids) with pixel-based visual streams. However, integrating pixel-based RGB features with 3D position information presents significant challenges. The two modalities exhibit divergent characteristics: 3D position data are spatially sparse and structured, while image features are spatially dense and unstructured. Directly concatenating or aligning these representations leads to feature incompatibility. To bridge this gap, we introduce a sinusoidal joint encoding strategy that projects 3D coordinates into a continuous sin-cos space. This transformation downscales noisy joint trajectories while preserving spatial relationships, enabling seamless fusion with visual features.

A second challenge arises from computational efficiency: processing full-frame video data at high temporal resolutions is prohibitively expensive for real-time systems, while naively fusing features across different temporal resolutions degrades performance. A common approach is to upsample low-frequency image features to match the motion sequence's time scale and merge them using a late fusion strategy~\cite{tani2024graph}. However, this final fusion step often results in feature loss due to the mismatch in the fineness of features, and it overlooks fine-grained temporal dependencies and misaligns crucial visual and motion cues. To address this, we propose a Multi-Modal Graph Convolutional Network (MMGCN) that processes high-frequency motion data via a hierarchical graph convolutional network and strategically integrates high-frequency motion features into low-frequency (1 fps) visual cues at a middle stage, ensuring a more coherent fusion with processed motion information later in the pipeline. By processing the two modalities in different temporal resolutions, our method effectively preserves fine-grained temporal dependencies and aligns visual motion cues efficiently. 

Finally, inspired by the observation that human actions transition smoothly over time, we develop SmoothLabelMix, a novel data augmentation technique that enforces temporal consistency by smoothing action labels along the temporal dimension and mixing input action sequences and labels in the spatial space.

 Overall, the technical contributions of the paper are:
	\begin{itemize}
		\item Sinusoidal encoding: A spatial encoding method that maps 3D skeleton coordinates into a continuous sin-cos space, enabling effective fusion with visual features.
		\item Multi-Modal Graph Convolutional Network: A hybrid graph-based architecture that processes 30 fps motion sequence through a graph convolutional backbone and integrates 1 fps visual data into skeleton sequences through a parallel branch, balancing computational efficiency and segmentation accuracy.
		\item SmoothLabelMix Augmentation: A data augmentation strategy combining label smoothing and weighted mixing to align inputs and labels bidirectionally. This reduces over-segmentation artifacts, improves generalization to ambiguous motion boundaries, and enhances robustness to noise.
	\end{itemize}
	
The remainder of this paper is organized as follows: Section~\ref{sec:2} reviews related work, Section~\ref{sec:3} details the Multi-Modal Graph Convolutional Network architecture, Section~\ref{sec:4} introduces the SmoothLabelMix data augmentation technique, Section~\ref{sec:5} presents 
experimental results, and Section~\ref{sec:6} concludes with future directions.

    
    \section{RELATED WORK}
\label{sec:2}
In this section, we briefly review three key areas related to our work: Human Action Segmentation, Multi-Modal Models, and Data Augmentation. 
\subsection{Human Action Segmentation}
Human Action Segmentation, which aims to temporally localize and classify sub-actions within continuous sequences, has evolved from early template-based methods~\cite{patron2010high, xu2012streaming, xing2021robust} to modern image-based learning architectures~\cite{lea2017temporal, farha2019ms}. Temporal Convolutional Networks (TCNs)~\cite{lea2017temporal} pioneered frame-wise prediction using dilated convolutions, while MS-TCN~\cite{farha2019ms} extended this with multi-stage refinement. Recent works like Action Segment Refinement Framework~\cite{ishikawa2021alleviating} combine boundary detection and classification to improve temporal localization. Despite progress, these methods suffer from high computational costs, sensitivity to background clutter, and the inclusion of redundant visual information, making them inefficient for real-time applications.

To address these challenges, graph-based networks are considered as a more efficient alternative by modeling human motion as structured spatial and temporal graphs, significantly reducing background noise and computational overhead. Early works, such as Spatial-Temporal Graph Convolutional Network (ST-GCN)~\cite{yan2018spatial}, introduced graph convolutional networks to sequentially capture joint-level motion patterns across both spatial and temporal dimensions. Subsequent improvements, like Two-Stream Adaptive Graph Convolutional Network (2s-AGCN)~\cite{shi2019two} and Multi-Scale Graph Network (MS-G3D)~\cite{liu2020disentangling}, incorporated attention mechanisms and multi-scale feature extraction to enhance action recognition. More recently, methods such as Pyramid Graph Convolutional Network (PGCN)~\cite{xing2022understanding} and Temporal Fusion Graph Convolutional Network (TFGCN)~\cite{xing2024understanding} have explored temporal modeling techniques to further improve segmentation accuracy.

However, despite these advancements, graph-based methods remain highly susceptible to noise from skeleton misdetections, leading to over-segmentation and fragmented action sequences. When combined with noisy object detections in human-object interaction scenarios, these errors further degrade segmentation quality, highlighting the need for robust multi-modal solutions.

\subsection{Multi-Modal Models} 
Multi-modal learning is a promising method for action recognition, leveraging complementary strengths of visual (RGB/depth) and skeletal modalities to overcome individual limitations. Early works, like~\cite{luo2014spatio, shahroudy2015multimodal} fuse images/depths and skeletons per frame
to improve the recognition results. Recent advancements focus on intermediate fusion strategies to better align cross-modal features. Kim \etal ~\cite{kim2023cross} proposed a cross-modal transformer that attends to spatial correspondences between skeleton joints and image regions, improving recognition consistency by dynamically weighting discriminative features. Similarly, Wang \etal ~\cite{wang2024cmf} integrates skeleton joints with RGB patches using a late fusion technique, but its reliance on dense pixel processing limits scalability. To address efficiency, PoseC3D~\cite{duan2022revisiting} encodes skeletons as pseudo-3D heatmaps and fuses them with RGB features via 3D convolutions, achieving a balance between accuracy and speed. More recently, Tani Hiroaki~\cite{tani2024graph} introduced a temporal attention mechanism that selects one representative image for the entire motion sequence capturing the minimal appearance features necessary for action recognition. While this approach is effective for short-term action recognition, it becomes computationally intensive for long-duration actions.

Building on advancements in multi-modal action recognition, our work extends these principles to human action segmentation, a task requiring precise temporal localization of sub-activities. While existing multi-modal models excel at classification, they often neglect the computational and temporal alignment challenges inherent to segmentation.

\begin{figure*}[ht]
    \centering
    \includegraphics[width=0.96\linewidth]{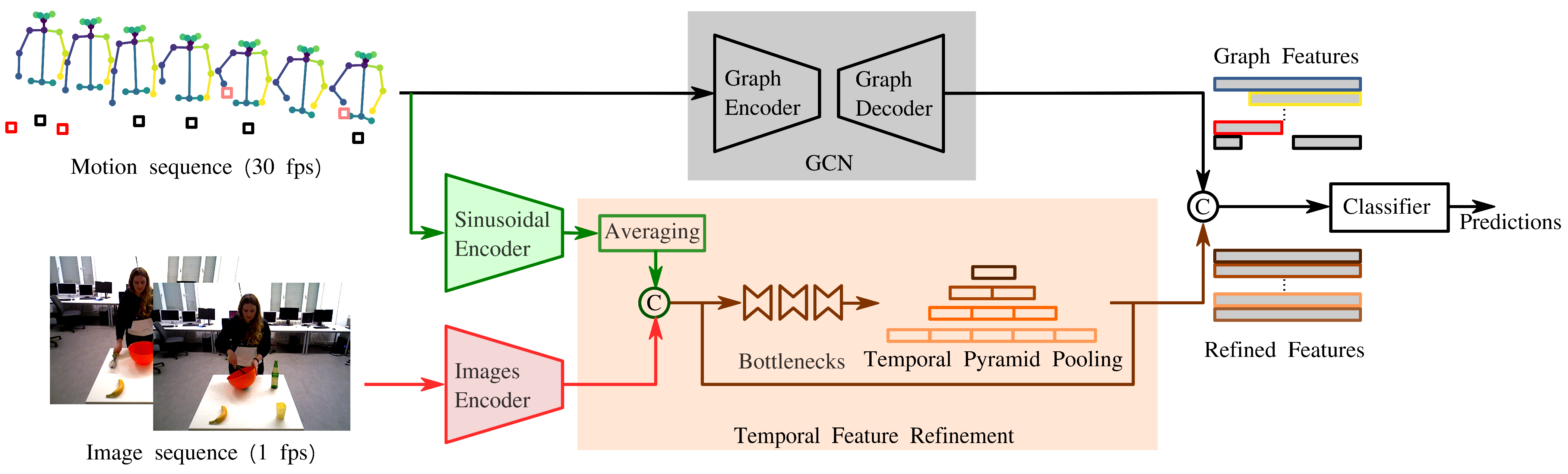}
    \caption{The framework of Multi-Modal Graph Convolutional Network with motion and image sequences in different temporal resolutions.}
    \label{fig:framework}
    \vspace*{-1.5\baselineskip}
\end{figure*}

\subsection{Data Augmentation}
Data augmentation plays a pivotal role in improving the generalization and robustness of temporal models, particularly for action recognition and segmentation tasks. Traditional techniques, such as spatial augmentations (e.g., cropping, flipping)~\cite{simonyan2014two} and temporal perturbations (e.g., frame skipping, time warping)~\cite{wang2016temporal}, focus on enhancing spatial invariance or simulating temporal variations. While effective for classification, these methods often disrupt the temporal coherence required for precise segmentation, where smooth transitions between sub-actions are critical.

Recent approaches adapt image-domain augmentation strategies to temporal tasks. Mixup~\cite{zhang2018mixup} and CutMix~\cite{yun2019cutmix}, which linearly interpolate inputs and labels, have been extended to videos by blending random clips (e.g., VideoMix~\cite{yun2020videomix}). However, these methods prioritize spatial or short-term temporal invariance, inadvertently introducing abrupt label transitions that exacerbate over-segmentation artifacts. For instance, TemporalMix~\cite{xiang2024joint} downsamples two segments by half, interpolates the larger segment to match the smaller one to preserve the natural smoothness of human actions, and then combines them into a new sequence. However, its downsampling process disrupts the natural tempo of human actions, leading to inconsistent predictions in speed-sensitive tasks.

To address these limitations, our work introduces SmoothLabelMix, a novel augmentation strategy tailored for action segmentation. Inspired by the observation that human actions transition gradually, SmoothLabelMix smooths the labels along the temporal dimension and blends action segments and their labels along the spatial dimension. This enforces a smooth label distribution that mirrors real-world action dynamics, while simultaneously training the model to handle noisy inputs caused by sensor errors or occlusions.

\section{Multi-Modal Graph Convolutional Network}
\label{sec:3}
The idea of the Multi-Modal Graph Convolutional Network is jointly optimizing dense spatial features from visual data and dense temporal dynamics from motion sequences (skeleton and object). Motion inputs (e.g., 30 fps) provide fine-grained motion patterns but suffer from joint tracking noise, while sparse RGB frames (1 fps) offer stable spatial context but lack temporal granularity. Our model bridges these modalities through two core mechanisms: a sinusoidal encoder and temporal feature refinement. 
\subsection{Sinusoidal Encoder}
To mitigate spatial jitter and enhance compatibility with visual features, we transform the joint position information in the camera coordinate into a continuous sin-cons representation. Given a 3D joint $P_i = (X_i, Y_i, Z_i)$, where $P_i\in \mathbb{R}^{1\times3}$, the encoder maps each coordinate into a high-dimensional space using sinusoidal and cosine functions. The encoder first scales these coordinates using a frequency factor determined by the parameters $\alpha$ and $\beta$. Then it computes sinusoidal $sin\_e$ and cosine $con\_e$ embeddings for each dimension, resulting in a high-dimensional vector that captures both the position and its spatial context. Mathematically, for each coordinate $c \in \{X_i, Y_i, Z_i\}$, the encoder computes:
\begin{align}
    sin\_e(c) &= sin(\frac{\beta\cdot c}{\alpha^{k/d}}), \\ 
    cos\_e(c) &= cos(\frac{\beta\cdot c}{\alpha^{k/d}}),
\end{align}
where $k$ is the index of the embedding dimension, and $d$ is the total number of dimensions. These sinusoidal and cosine embeddings are concatenated to form the final encoded representation $P_e$:
\begin{equation}
    P_{e, i} = concat(sin\_e(P_i), cos\_e(P_i)).
\end{equation}

We set $d=128$ and obtain a $768$ dimensional embedded feature map ($3$ coordinates $\times 128 \times2$ functions). The $\alpha$ is set as $1000$ to create geometrically increasing wavelengths from $1$ to $1000$ units, capturing both fine and coarse-grained spatial features. The $\beta=100$ scales the values of the real-world coordinates. A $1\times1$ convolution then compresses this to a target dimension ($512$ channels), balancing expressiveness and efficiency.

While transformer-based models like GCNs employ learned positional encodings to adaptively update inter-node relations through attention mechanisms, this flexibility comes with significant computational costs. In contrast, the sinusoidal encoder provides a fixed geometric prior that explicitly preserves spatial relations through wavelength-scaled trigonometric function. The fixed encoding prevents overfitting to sparse training skeletons while maintaining spatial continuity between positions. This proves particularly valuable when aligning with sparse visual features where spatial precision matters more than temporal sequencing. More evidence can be found in the ablation studies in the Section~\ref{sec:5}.


\subsection{Temporal Feature Refinement}
While position information is encoded using the sinusoidal encoder described earlier, the image information is processed into high-dimensional features through a modified I3D model~\cite{carreira2017quo}, in which the stride along the time axis is removed because of the limited frames. By concatenating the motion and image features together, a multi-modal feature is formed. The Temporal Feature Refinement module is designed to integrate these multi-modal features into a unified representation and upsample them to the time scale of the graph motion sequence. 


As demonstrated in Fig.~\ref{fig:framework}, the module begins by averaging the encoded position features to match the temporal-spatial resolution of the image features. Then the concatenated features are refined by three consecutive bottleneck layers, where the features are compressed in half at the bottleneck place and then remapped back to the high dimensions. The refined features are further upsampled to the motion time scale through a temporal pyramid pooling layer~\cite{xing2022understanding} with four parallel temporal pooling blocks. The final refined features are obtained by introducing the interpolated original features into the pooled features.

\subsection{Multi-Modal Graph Convolutional Network}
In addition to the motion and image feature refinement stream, the original motion sequence is processed in a graph encoder-decoder network~\cite{xing2024understanding}. This network models the structural dependencies between motion elements (skeleton and objects) by representing them as spatial and temporal graphs. The graph nodes are skeleton keypoints and object center points, and the graph edges represent the dynamic relations between nodes. Through multiple layers of graph convolutions and pooling operations, the network learns high-dimensional motion features that preserve both local and global motion contexts. The output of the graph encoder-decoder has the same dimensionality as the motion feature refinement stream, allowing for seamless fusion.

The extracted graph-based motion features and the refined motion-image features are concatenated and passed through a multi-layer classifier. This classification head leverages two cascaded 2D convolutional kernels to complement the strengths of both streams to improve recognition performance. As shown in Figure~\ref{fig:framework}, the complete MMGCN framework integrates the following modules: Sinusoidal Encoder, Image Encoder, Temporal Feature Refinement, Graph Convolutional Network, and a Classifier.

\section{Smooth Label Mix}
\label{sec:4}

Human motion smoothly transitions from one action to the next, whereas action labels change in binary form from 0 to 1 or vice versa. To bridge this discrepancy, we propose SmoothLabelMix, a two-stage framework for training sequence models with smoothed labels and intra-batch mixing. The method consists of two key components: (1) \textbf{Label Smoothing} using linear or Gaussian filters, and (2) \textbf{Weighted Mixing} of input sequences and their corresponding smoothed labels.

\begin{figure}[t]
    \centering
    \includegraphics[width=0.9\linewidth]{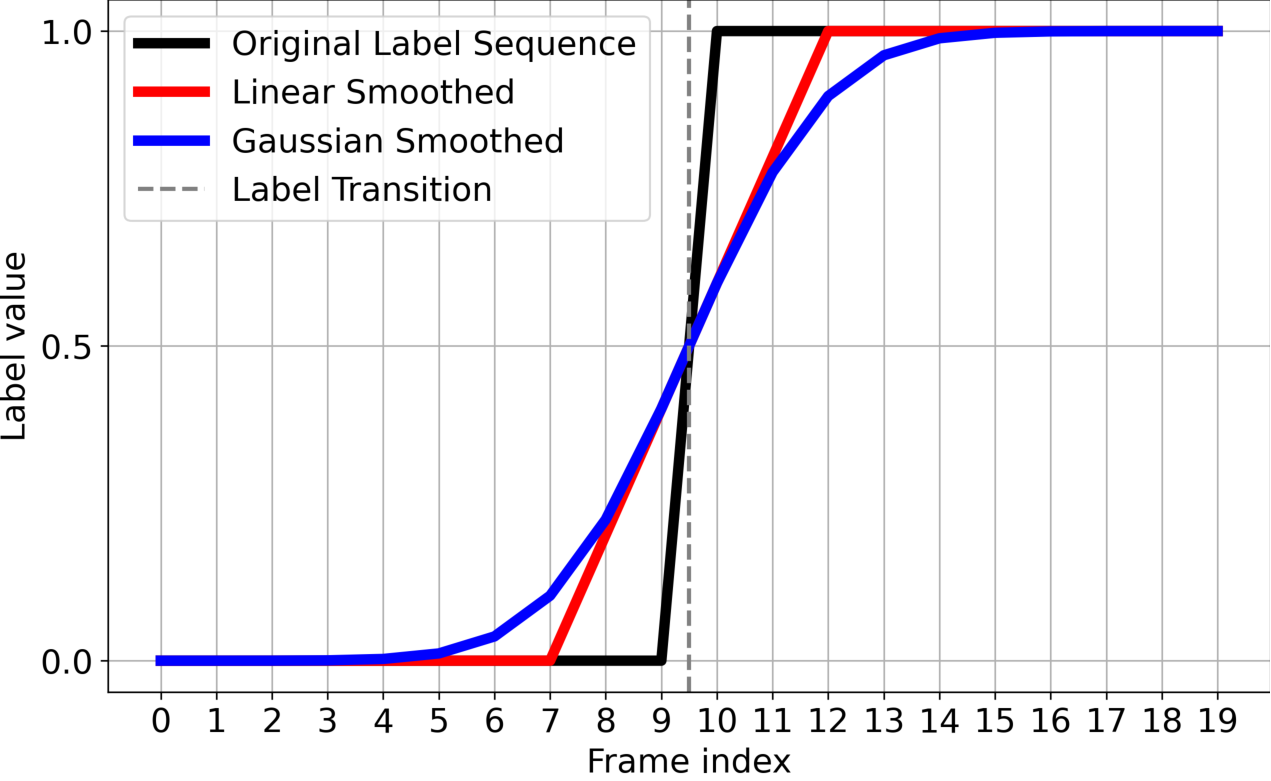}
    \caption{Comparison of Original, Linear, and Gaussian Smoothed label sequences, all sequences have a label transition at the same frame index.}
    \label{fig:smooth_label}
\end{figure}

The Label Smoothing technique aligns the label transitions with the smooth nature of human motion. Specifically, we implement two types of filters (linear and Gaussian) to process the binary label sequences, as illustrated in Figure~\ref{fig:smooth_label}. These filters create gradual transitions in the labels, better reflecting the continuous nature of human motion. The linear filter applies a uniform averaging over a fixed window, resulting in piecewise-linear transitions, while the Gaussian filter uses a weighted average based on the Gaussian distribution, producing smoother, more natural S-curve transitions. By incorporating these smoothed labels, we aim to improve the alignment between the model's predictions and the inherent continuity of human actions.

The {Weighted Mixing} technique further enhances model robustness by combining pairs of input sequences and their corresponding smoothed labels within a training batch. Given two input sequences $x_1$ and $x_2$, the mixed sequence $x_{mix}$ is computed as:
\begin{equation}
    x_{mix} = w\cdot x_{1} + (1-w)\cdot x_{2}
\end{equation}
where $w$ is a weighting factor with $w\sim Beta(\alpha=0.2)$, balances the contribution of each input. Similarly, the corresponding labels are mixed using the same weighting scheme to maintain consistency between the inputs and annotations. This method enhances the model’s ability to generalize across gradual action transitions, reducing abrupt segmentation errors and improving temporal coherence.

    \section{EXPERIMENTS AND RESULTS}
\label{sec:5}
To evaluate the performance of the proposed modules, we conduct the experiments on a public human-objects interaction dataset: Bimanual Actions~\cite{dreher2019learning}. On this dataset, we also perform systematic ablation studies to validate four critical design choices:
\begin{enumerate}
    \item Sinusoidal encoding: ablation with/without temporal positional embeddings.
    \item Fusion strategies: early fusion (input-level concatenation), middle fusion (feature-level attention), and late fusion (output-level averaging).
    \item Label filters: linear vs. Gaussian filtered label transitions.
    \item Weighted mixing: ablation with/without mixing data.
\end{enumerate}

\subsection{Datasets}

\begin{table*}[t!]
  \caption{Ablation studies: the accuracy of framewise prediction and F1@k score of action segmentation using models with different modifications in each unit$^a$}
  \label{tab:ablation_study}
  \centering
  \begin{tabular}{c c c| c c | c c | c c | c c c | c c c c c}
    \toprule
    \multicolumn{3}{c|}{Smoothing$^b$}& \multicolumn{2}{c|}{Mixing}  & \multicolumn{2}{c|}{Refinement}& \multicolumn{2}{c|}{Sinusoidal} &  \multicolumn{3}{c|}{Fusion Strategies $^c$} &
    \multicolumn{5}{c}{Evaluation Metrics ($\%$) $^d$}
    \\         

    O & L & G & w & w/o & w & w/o & w & w/o & Early & Mid & Late & Top 1  & F1 macro & F1@10 & F1@25 & F1@50\\
    \midrule
        $\times$  &  &  & & $\times$ &  & $\times$ & & $\times$  &  & &  &
        82.91& 81.93&91.14 &88.39 & 78.26\\ 

        & $\times$ &  & & $\times$ &  & $\times$ & & $\times$ &  & &  &
        83.83& 82.82& 92.03&89.14 &77.30 \\

        &  &  $\times$  &  &  $\times$ &   & $\times$ & & $\times$ & & & &
        84.05&83.77 &90.63 & 88.44&77.85 \\
        $\times$  &  &  & $\times$&  &   & $\times$ & & $\times$ &  & &  &
        84.70& 84.68&91.33 & 89.15&80.61 \\
        & $\times$ &  & $\times$&  &   & $\times$  &  &$\times$  &  & &  &
        \underline{89.09} & \underline{89.31} & 93.41 & 91.59& 80.65\\
        & & $\times$  & $\times$ &  &  & $\times$  &  &$\times$  & &  & & 
        88.92&88.89 &\underline{94.14} &\underline{92.52} & $\underline{\boldsymbol{84.07}}$\\ 
        \midrule
        & & $\times$ & $\times$ & & & $\times$ & &$\times$  & & & $\times$ 
        &88.48& 88.65&92.88 & 90.95 & 81.49  \\ 

        & & $\times$ & $\times$ & &$\times$ & & &$\times$  & & & $\times$ 
        &89.71& 89.89 & 93.80 & 91.77 & 83.49  \\
        & & $\times$ & $\times$ & & $\times$ & & & $\times$ & & $\times$ &  &
        89.74 & 89.95 & 94.81 & 92.71 & 83.64 \\
        &  & $\times$ & $\times$ &  &$\times$ & & &$\times$ & $\times$ & & 
        &88.42 &88.53 & 92.31& 89.82 &80.48\\

        &  & $\times$  & $\times$ &  & $\times$ &  &$\times$ & & & $\times$ 
        &$\times$ &
        $\boldsymbol{90.19}$ & $\boldsymbol{90.39}$ & $\boldsymbol{95.26}$ & $\boldsymbol{93.05}$ & 83.87 \\

    \bottomrule
  \end{tabular}
  \scriptsize
  \begin{tablenotes}
    \item[a]$^a$ ``Smoothing" means Label Smoothing, ``Mixing" represents Weighted Mixing, ``Sinusoidal" is the Sinusoidal Encoder, and ``Refinement" is the Temporal Feature Refinement block.
    \item[a]$^b$ ``O" is the original label, ``L" means the linear filtered label, and ``G" means the Gaussian filtered label. ``w" is with and ``w/o" means without.
    \item[a]$^c$ Empty entries indicate models trained solely on motion sequences without visual input.
    \item[a]$^d$ The best results comparing all modifications are in \textbf{bold}; The best results among different configurations of label smoothing and weighted mixing are \underline{underlined}.
  \end{tablenotes}
\end{table*}

\textbf{Bimanual Actions dataset}~\cite{dreher2019learning} captures fine-grained human-object interactions for kitchen tasks comprising 540 recordings (2 hours 18 minutes total) of bimanual manipulation tasks such as cutting and stirring. This dataset provides comprehensive framewise annotations including 3D bounding boxes for 12 objects, 3D skeletons for 6 human subjects and 14 interaction categories (including both-hand interactions). For our experiments, we represent each 3D skeleton using 15 upper-body joints (5 head joints, 6 arm joints, and 4 torso/hip joints) and represent objects by their 3D bounding box centroids. Following the dataset authors' protocol, we employ \textit{leave-one-subject-out} cross-validation to evaluate generalization to unseen users. For ablation studies, we specifically evaluate on Subject 1's test set to isolate the impact of proposed components (e.g., label smoothing, fusion strategies), while comparing against state-of-the-art methods on the full benchmark.


For our multi-modal fusion experiments, each input sample integrates a 120-frame motion sequence (comprising 3D body skeletons and object bounding box center points) with $4$ RGB images ($480\times 640$) uniformly sampled from the video clip. This configuration creates a 30:1 time resolution ratio between the high-frequency motion data (120 frames) and the sparsely sampled visual data (4 frames)

\subsection{Experimental Settings}

We evaluate our method using standard metrics to ensure a comprehensive assessment of its performance. For evaluation, we employ metrics such as recognition accuracy, micro-F1, and segmentation accuracy (F1@k). The model is optimized using the SGD optimizer with an initial learning rate of $10^{-4}$, weight decay of $0.0005$, and multi-step learning rate decay at epochs 20 and 40. Training runs for 60 epochs on two NVIDIA RTX 2080ti GPUs, leveraging mixed precision to enhance efficiency and reduce memory overhead. We apply the SmoothLabelMix data augmentation techniques to improve generalization. The evaluation experiments are conducted on a single GPU.

\begin{figure}[t]
    \centering
    \begin{tabular}{@{}ccc@{}}
    \includegraphics[width=0.31\linewidth]{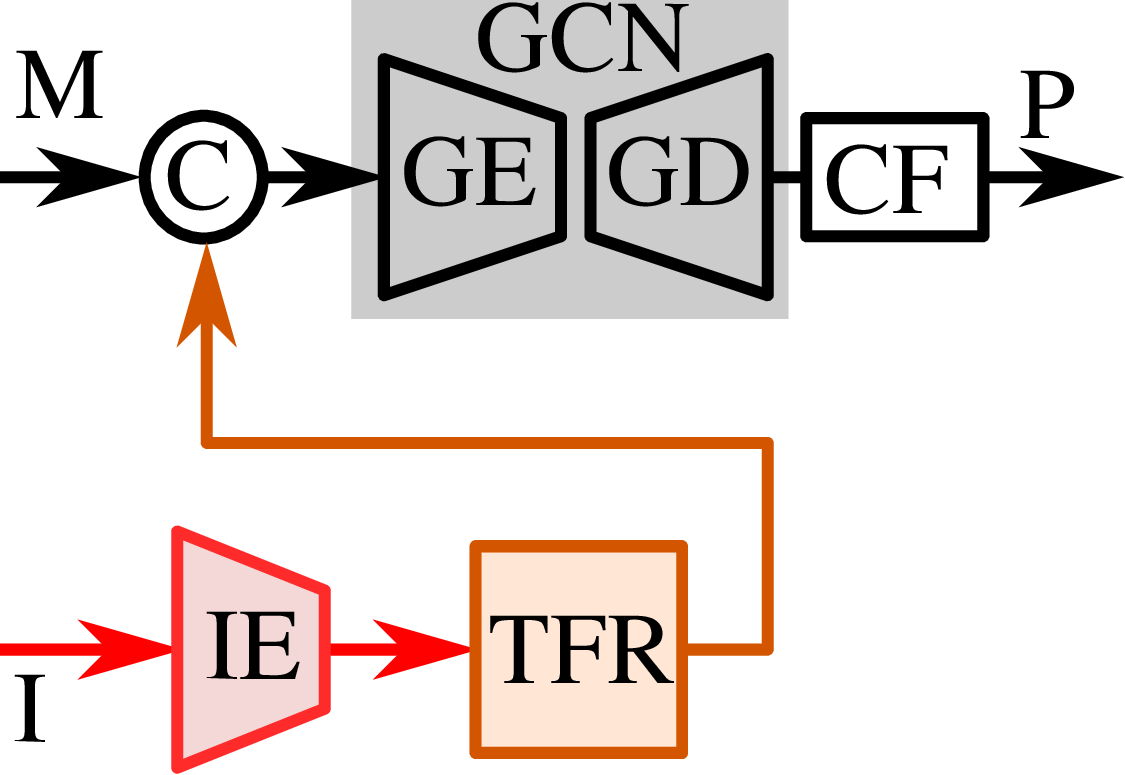} &
    \includegraphics[width=0.29\linewidth]{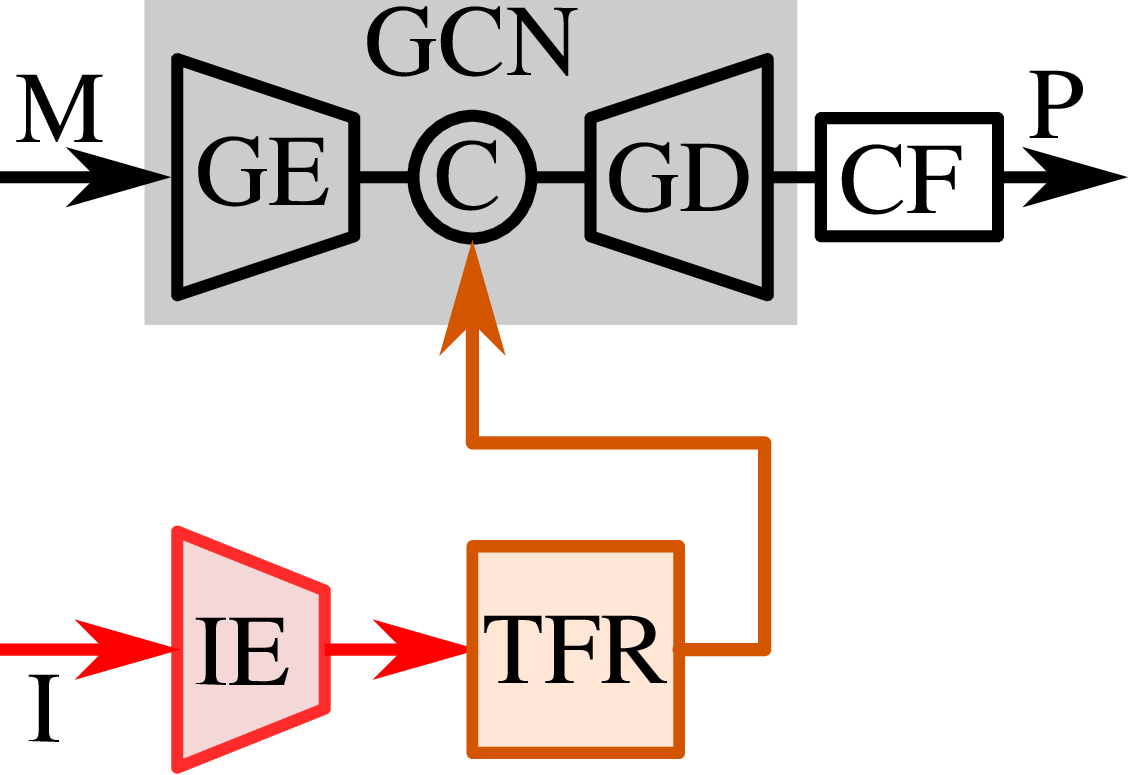} & 
    \includegraphics[width=0.3\linewidth]{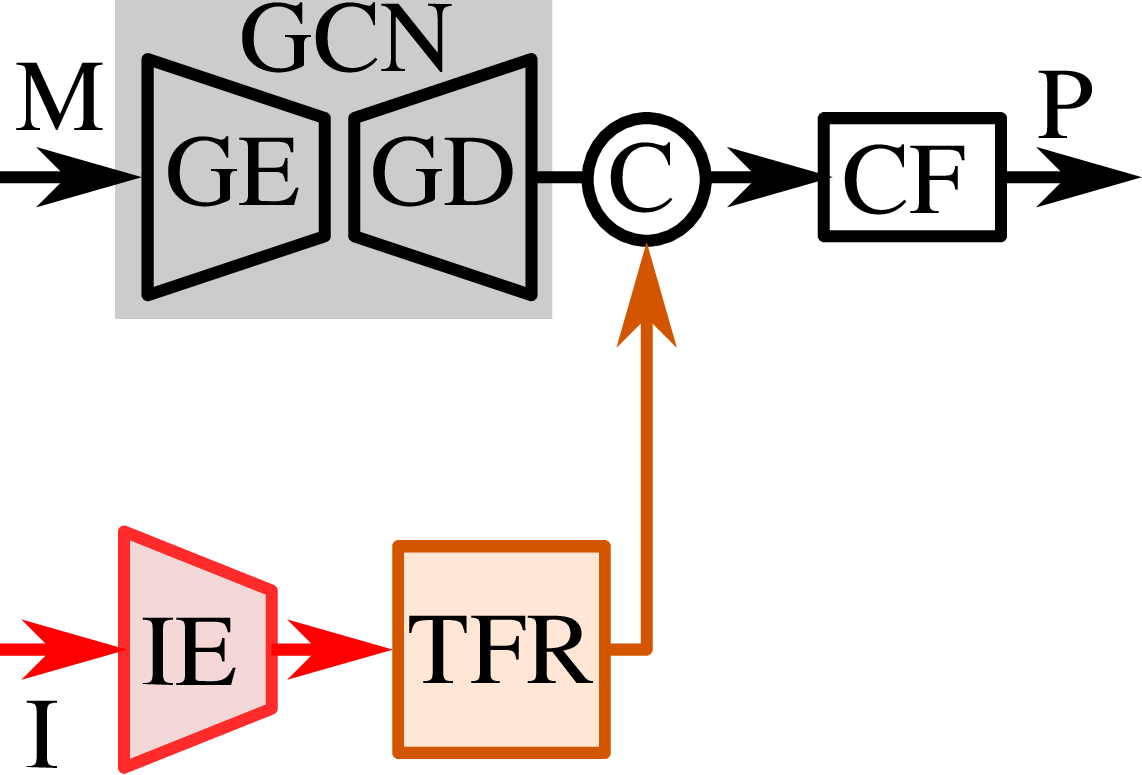} \\
    (a) & (b) & (c)
    \end{tabular}
  \caption{\label{fig:fusion_strategies} Three fusion strategies with the Temporal Feature Refinement module: (a) early fusion, (b) middle fusion, (c) later fusion. I: image sequence, M: motion sequence, IE: image encoder, TFR: temporal feature refinement, GE: graph encoder, GD: graph decoder, CF: classifier, P: predictions.}
\end{figure}

\subsection{Ablation Study}
Our ablation study systematically evaluates the contributions of three core innovations in the framework: SmoothLabelMix data augmentation, the Temporal Refinement module, and the Sinusoidal Encoder within the multi-modal graph convolutional network. The introduction of the image naturally raises the challenge of fusion strategies. Therefore, while analyzing the Sinusoidal Encoder, we also explored the impact of different fusion approaches: early, middle, late, and middle+late. Figure~\ref{fig:fusion_strategies} illustrates three representative examples of these strategies integrated with the \textit{Temporal Feature Refinement} block. These strategies differ primarily in the stage at which image and motion features are fused into the main processing stream. For configurations omitting the \textit{Temporal Feature Refinement} module, the encoded features are upsampled via a linear interpolation layer instead. Note that we categorize the implementation of the \textit{Sinusoidal Encoder} as a middle fusion method, because the \textit{Sinusoidal Encoder} processes the motion features before fusion within the Temporal Refinement.

\begin{table*}[t]
  \caption{Comparison of cross-validation results with state-of-the-art methods on the Bimanual Actions dataset~\cite{dreher2019learning}$^a$}\label{tab:sota_bimacts}
  \centering
\footnotesize
  \begin{tabular}{p{0.2\textwidth} p{0.1\textwidth} p{0.1\textwidth} p{0.1\textwidth} p{0.1\textwidth} p{0.1\textwidth}}
    \toprule
      Model & Accuracy ($\%$) & F1 macro ($\%$) & F1@10 ($\%$) & F1@25 ($\%$) & F1@50 ($\%$) \\
    \midrule
      Dreher et al. \cite{dreher2019learning} & $63.0$ & $64.0$ & $40.6 \pm 7.2$ & $34.8 \pm 7.1$ & $22.2 \pm 5.7$ \\
      H2O+RGCN~\cite{lagamtzis2023exploiting} & 68.0 & 66.0 & $-$ & $-$ & $-$ \\
      Independent BiRNN \cite{morais2021learning} & $74.8$ & $76.7$ & $74.8 \pm 7.0$ & $72.0 \pm 7.0$ & $61.8 \pm 7.3$ \\
      Relational BiRNN \cite{morais2021learning} & $77.5$ & $80.3$ & $77.7 \pm 3.9$ & $75.0 \pm 4.2$ & $64.8 \pm 5.3$  \\
      ASSIGN~\cite{morais2021learning} & $82.6$ & $79.8$ & $84.0 \pm 2.0$ & $81.2 \pm {2.0}$ & $68.5 \pm {3.3}$  \\
      2G-GCN~\cite{qiao2022geometric} & $-$ & $-$ & $85.0\pm 2.2$ & $82.0 \pm 2.6$ & $69.2 \pm 3.1$ \\
      PGCN~\cite{xing2022understanding} & ${86.8}$ & ${83.9}$ & ${88.5}\pm {\bm{1.1}}$ &${85.5} \pm {2.0}$ & ${77.0} \pm 3.4$\\
      {TFGCN}~\cite{xing2024understanding} & ${88.4}$ & ${88.6}$ &${93.7} \pm 1.2$ & ${91.9} \pm{\mathbf{1.6}}$ & ${85.4} \pm {\mathbf{2.9}}$  \\
      \textbf{MMGCN} (ours) & $\mathbf{89.3}$ & $\mathbf{89.3}$ & $\mathbf{94.5}\pm 1.6$ & $\mathbf{92.8}\pm{1.9}$ & $\mathbf{86.1}\pm {3.1}$  \\
    \bottomrule
  \end{tabular}
  \scriptsize
  \begin{tablenotes}
        \item[a]$^a$ The models are cross validated on the leave-one-subject-out benchmark, the best results of each class are in \textbf{bold}. The Accuracy and F1 micro results are averaged, 
        and F1@k are listed with mean and standard deviation. 
        A smaller standard deviation value indicates greater robustness in the model.
  \end{tablenotes}
\end{table*}

The first $6$ rows of the ablation study on Table~\ref{tab:ablation_study} reveal critical insights into the interplay between label smoothing and data mixing for action recognition and segmentation. Considering the Top1 and micro F1 performance in action recognition, combining label smoothing with data mixing consistently outperforms non-mixed counterparts, with the largest gains observed for linear smoothing + mixing ($89.09/89.31$ vs $83.83/82.82$ without) and Gaussian smoothing + mixing ($88.92/88.89$ vs $84.05/83.77$ without). This demonstrates that mixing enhances model robustness by enforcing consistency between smoothed labels and motion inputs. Considering the F1@k performance,  the benefits of Gaussian smoothing + mixing are most pronounced at strict temporal thresholds (F1@50: $84.07$ vs. $77.85$ without mixing), indicating better handling of action boundaries. Notably, segmentation at loose thresholds (F1@10/25) improves across all configurations, suggesting label smoothing alone helps with coarse temporal alignment, while mixing refines fine-grained transitions. The synergy between Gaussian smoothing and data mixing achieves the best segmentation performance across the configurations of smoothing and mixing techniques, reducing over-segmentation by $5.81\%$ (F1@50) compared to the baseline (the first row). Given its balanced improvement across recognition accuracy and segmentation precision, particularly at challenging temporal thresholds, we adopt the Gaussian label smoothing and data mixing combination for all subsequent experiments.

Beginning at the $7$-th row, we introduce the image stream alongside the motion sequence as multi-modal inputs. Rows 7-8 of the ablation study (Table~\ref{tab:ablation_study}) evaluate the impact of the proposed Temporal Refinement module, which enhances the integration of visual and motion modalities. Integrating the Temporal Refinement module improves the frame recognition accuracy by $1.23\%$, while also mitigating over-segmentation, particularly beneficial for F1@50 ($+2.00\%$). The results demonstrate the module's effectiveness in enhancing temporal coherence and segmentation precision. However, the inclusion of image streams still brings disturbance to the model, especially for action segmentation. This observation motivates us to further analyze the configuration of the fusion strategy and the Sinusoidal Encoder.

Rows 8-11 of the ablation study (Table~\ref{tab:ablation_study}) reveal critical insights into the efficacy of the fusion strategies and the Sinusoidal Encoder. Early fusion alone performs poorly (Accuracy: 88.42, F1@50: 80.48), as raw skeleton and image features suffer from noise amplification and temporal misalignment. Integrating the Sinusoidal Encoder into a late fusion significantly mitigates these issues, which projects 3D skeleton coordinates into a continuous sin-cos space. The combination of Sinusoidal (middle) and late fusion strategy achieves the best performance (Accuracy: $90.19\%$, F1@10: $95.26\%$), demonstrating that sinusoidal encoding stabilizes pose features by smoothing joint trajectories, and late fusion refines boundaries using contextual visual cues. These results underscore the necessity of the Sinusoidal Encoder for the superiority of decoupling early noise suppression from late contextual refinement in multi-modal frameworks. 

The middle, and late fusion variants all fuse sparse appearance features with motion representations processed through a transformer-based encoder (Graph Encoder). The middle+late fusion substitutes this transformer encoder with our \textit{Sinusoidal Encoder}, which operates as a middle fusion component by processing raw coordinates before temporal refinement. This architectural distinction proves that the \textit{Sinusoidal Encoder}'s fixed geometric priors outperform learned transformer embeddings in noise suppression while maintaining spatial coherence. It enables a more effective alignment with sparse visual features.

For efficiency analysis, our model archives an optimal balance between computational efficiency and performance, operating at $131.0$ GFLOPs, significantly lower than using higher visual temporal resolution. For instance: processing motion and visual modalities at a resolution ratio of $30:2$ increases FLOPs to $220.0$ G,
while uniform high-resolution processing ($30:30$) escalates costs to $2687.3$ GFLOPs, a $20.5\times$  increase. This exponential scaling highlights the inefficiency of naively aligning modalities at high frequencies. While our framework introduces moderate computational overhead compared to models using pure motion sequences, it retains real-time performance (81.0 FPS on a single GPU) by strategically decoupling temporal resolutions.

\subsection{Comparison with states-of-the-art}

\begin{figure*}
    \centering
    \includegraphics[width=0.99\linewidth]{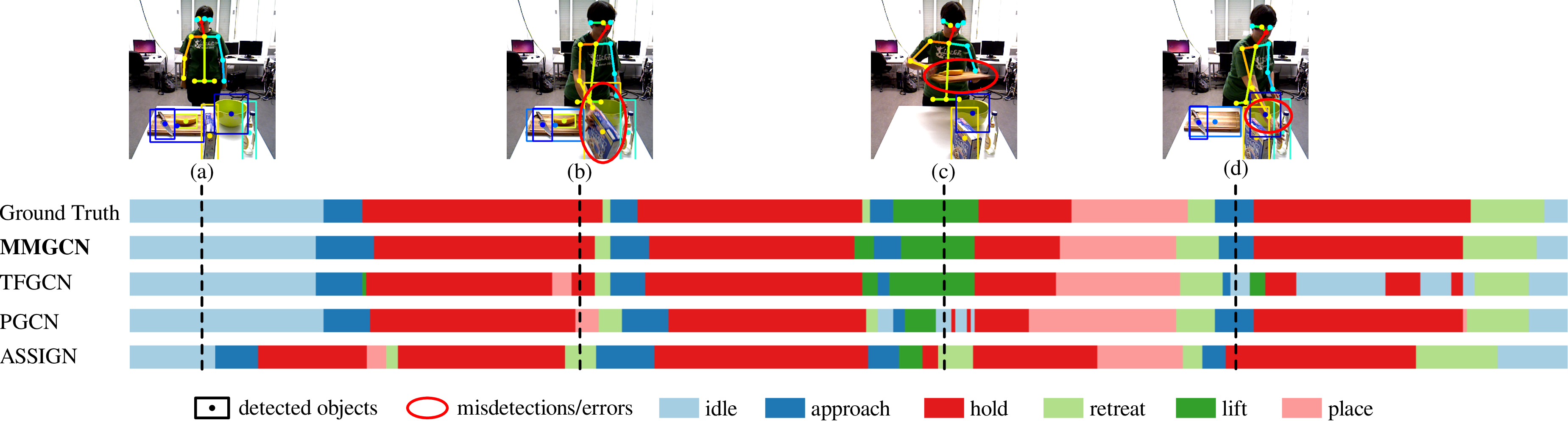}
    \caption{Qualitative comparison of action segmentation results: Visualized action segmentation outputs of different models for the left arm in the breakfast cereal preparation scenario~\cite{dreher2019learning}, illustrating the temporal alignment and accuracy of predicted action boundaries. Top: input frames with perception artifacts (a-d). Bottom: ground-truth and predictions.}
    \label{fig:qualitative}
\end{figure*}

\begin{figure}[t]
    \centering
    \begin{tabular}{@{}cc@{}}
    \includegraphics[width=0.22\textwidth]{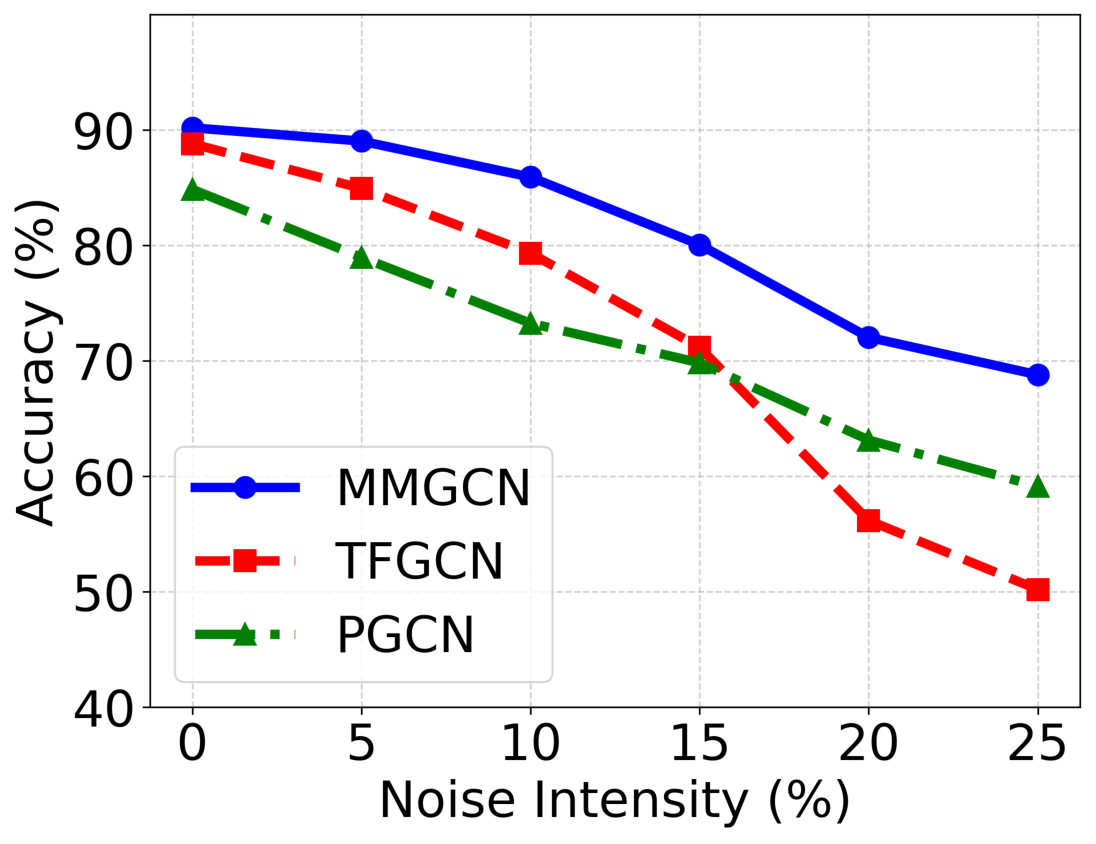} &
    \includegraphics[width=0.22\textwidth]{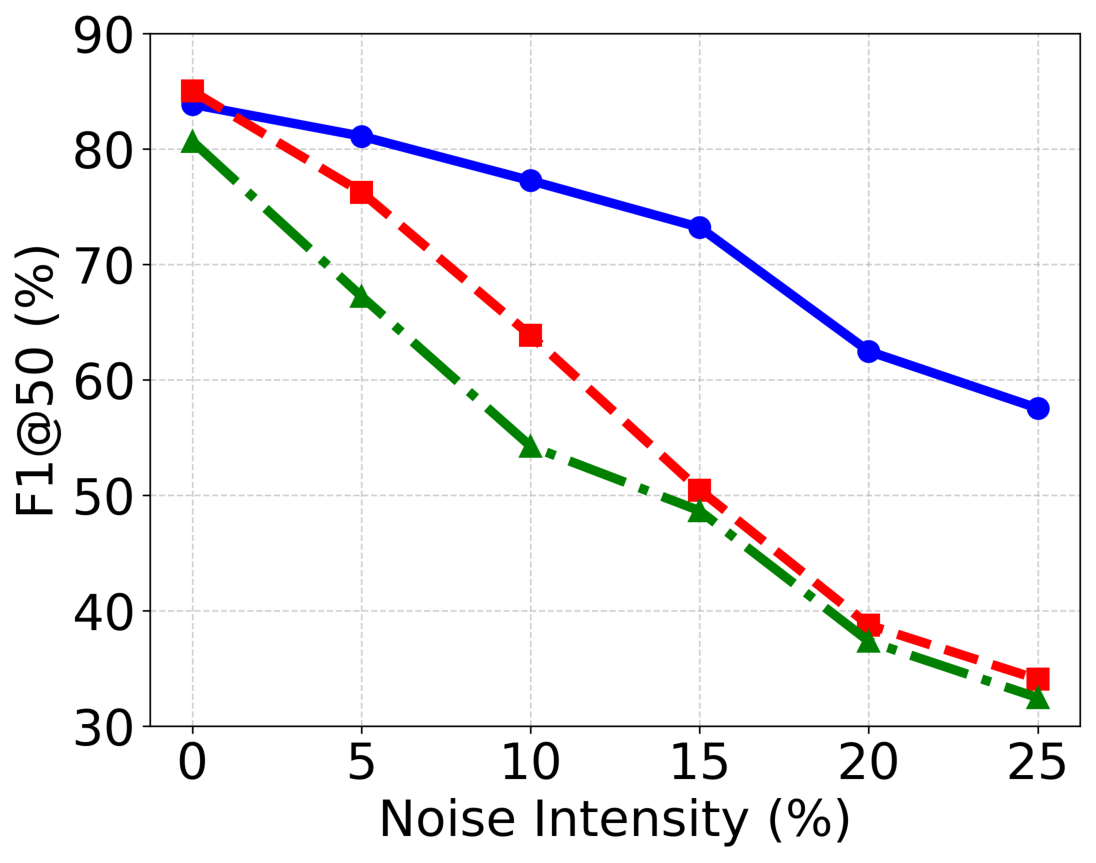} \\
    (a) & (b)
    \end{tabular}
  \caption{\label{fig:robustness} Comparison of the robustness of three models under increasing noise intensities. The figures illustrate how (a) action recognition accuracy and (b) segmentation F1@50 vary as noise levels rise.}
\end{figure}

The proposed MMGCN is compared with state-of-the-art methods in the field of human action segmentation on the Bimanual Actions dataset~\cite{dreher2019learning}. The compared methods include the method proposed by Dreher \etal~\cite{dreher2019learning}, H2O+RGCN~\cite{lagamtzis2023exploiting}, BiRNN~\cite{morais2021learning}, 2G-GCN~\cite{qiao2022geometric}, PGCN~\cite{xing2022understanding} and TFGCN~\cite{xing2024understanding}.

The cross-validation results are listed in Table~\ref{tab:sota_bimacts}. Our proposed MMGCN achieves state-of-the-art performance across all evaluation metrics, as validated on the leave-one-subject-out benchmark. Compared to TFGCN~\cite{xing2024understanding}, the previous best method, MMGCN improves accuracy from $88.4\%$ to $89.2\%$ and F1 macro from $88.6\%$ to $89.1\%$, with significant gains in boundary-sensitive metrics: F1@10 increases from $93.7$ to $94.3$, F1@25 from $91.9$ to $92.7$, and F1@50 from $85.4$ to $85.9$. Compared to PGCN~\cite{xing2022understanding}, which achieved $88.5\%$, $85.5\%$, and $77.0\%$ action segmentation performance, MMGCN demonstrates substantial improvements, particularly at higher overlap thresholds, indicating better temporal localization of action boundaries. These improvements highlight MMGCN’s robustness to temporal misalignment and the efficacy of the multi-modal fusion strategy. Despite achieving superior performance, MMGCN exhibits a slightly higher standard deviation compared to TFGCN~\cite{xing2024understanding}, particularly at the F1@10 and F1@25. This increased variability might be attributed to differences in model architecture and sensitivity to variations in motion and image inputs. This observation motivates further experiments under noise-induced conditions to evaluate the model’s robustness, ensuring consistent performance across diverse scenarios.

The observed average improvement of $1\%$ in Top-1 accuracy compared to state-of-the-art methods motivates further analysis of our approach's robustness in real-world scenarios. To simulate realistic conditions, we progressively introduce input noise (simulating detection failures) by randomly removing motion nodes from both human joints and object centroids at each frame in Subject 1's test set from the Bimanual Actions dataset~\cite{dreher2019learning}. We systematically increase the noise intensity (node removal rate) from $0\%$ to $25\%$. For each frame, we randomly select $V' = \eta \times V$ nodes without replacement (where $\eta$ is the noise intensity ranging 0-$25\%$), setting all feature channels of selected nodes to zero. The results of robustness are illustrated in Fig.~\ref{fig:robustness}. As noise intensity increases, the gap in action (a) recognition and (b) segmentation accuracy between the MMGCN and the other two models progressively widens. Notably, the MMGCN consistently maintains the highest performance across all noise levels and, in the segmentation case, overtakes competing models to secure the top position. This trend highlights the superior resilience of our approach to partially missing inputs. 

Figure~\ref{fig:qualitative} presents the qualitative comparison of action segmentation results for the breakfast cereal preparation scenario from the Bimanual Actions dataset~\cite{dreher2019learning} with input frames illustrating critical perception artifacts. Frame (a) show correct detections, frames (b-c) exhibit object misdetections (missing bowl, cutting board), and frame (d) contains depth errors from hand-object occlusion (miscalculated bowl position). While comparative models (TFGCN, PGCN, ASSIGN) exhibit severe over-segmentation due to these input artifacts, particularly misaligning "hold", "lift" and "approach", our \textit{MMGCN} successfully mitigates the over-segmentation issues and maintains coherent segments closest to ground truth. This demonstrates the model’s effectiveness in maintaining coherent action segments. However, MMGCN overlooks actions with short durations or extends their intervals, leading to temporal shifts in action boundaries. 
	
    \section{CONCLUSIONS}
\label{sec:6}

In this study, we propose the Multi-Modal Graph Convolutional Network (MMGCN), a novel framework for robust human action segmentation that harmonizes high-frequency motion data (e.g., 30 fps) with low-frequency visual cues (e.g., 1 fps) via a Sinusoidal Encoder and a mid-stage fusion strategy. The Sinusoidal Encoder projects 3D joint coordinates into a continuous spatial-temporal embedding, while our hybrid fusion architecture integrates motion and visual features at divergent temporal resolutions, preserving fine-grained dependencies and computation efficiency. Complemented by SmoothLabelMix, a data augmentation technique that synthesizes realistic motion-noise transitions, MMGCN achieves state-of-the-art performance across the action recognition and segmentation benchmark.


The model's sensitivity to variations in input data motivates further research into enhancing its robustness. Future work will focus on investigating the model's performance under noisy and incomplete input conditions by incorporating noise injection techniques during training. Additionally, we plan to explore alternative fusion mechanisms, advanced attention models, and extended benchmarks on diverse datasets to validate the model's generalizability and practical applicability across different action recognition tasks.


    \section*{ACKNOWLEDGMENT}
The research has been (partially) supported by the German Research Foundation (DFG) SFB 1320 EASE, CRC, University of Bremen. The research was conducted in subproject R1: NEEM-based embodied knowledge system. We extend our sincere gratitude to Timothy Ju Kin, Ng for his invaluable technical insights and support.
	

	
	
	\bibliographystyle{IEEEtran}
	\bibliography{IEEEexample}

\end{document}